\documentclass{article}

\usepackage[T1]{fontenc}
\usepackage{appendix}
\usepackage{amssymb}
\usepackage{microtype}
\usepackage{booktabs}
\usepackage{array}
\usepackage{amsmath}
\usepackage{graphicx}
\usepackage{caption}
\usepackage{multirow}
\usepackage{hyperref}
\usepackage{xurl}

\title{Ensuring Reliability in Programming Knowledge Tracing: A Re-evaluation of Attention-augmented Models and Experimental Protocols}

\author{
Jaewook Kim \\
Korea University, Seoul, Republic of Korea \\
\texttt{lesit@korea.ac.kr}
\and
Hyeoncheol Kim\thanks{Corresponding author} \\
Korea University, Seoul, Republic of Korea \\
\texttt{harrykim@korea.ac.kr}
}

\date{}

\begin{document}

\noindent
{\small
This is a preprint of a paper accepted at the International Conference on Intelligent Tutoring Systems (ITS 2026), to appear in Springer LNCS.
}

\vspace{0.3em}
\noindent\rule{\textwidth}{0.4pt}
\vspace{0.5em}

\begingroup
\let\newpage\relax
\maketitle              
\endgroup

\begin{abstract}
Programming Knowledge Tracing (PKT) has recently advanced through hybrid approaches that integrate attention-based feature modeling for code representation with RNN-based sequential prediction. While these models report strong empirical performance, their reliability can be sensitive to subtle implementation and experimental design choices. This study revisits representative PKT models and shows that reported gains can be substantially influenced by model configuration and sequence construction practices. We identify issues in attention dimension settings that affect performance estimates, and demonstrate that improper ordering of student attempts—such as ignoring ServerTimestamp—can violate temporal causality and lead to overly optimistic results. To ensure consistent evaluation, hyperparameters are selected via grid search guided by a single designated fold and then fixed uniformly across all folds during cross-validation. We further analyze the role of assignment-wise characteristics and systematically explore the impact of maximum sequence length. Using this protocol, we re-evaluate PKT models on the CodeWorkout dataset. Our results show that, under controlled and consistent settings, the performance gap between attention-enhanced models and standard DKT is significantly reduced, and increased architectural complexity does not consistently translate into superior performance. Beyond individual model comparisons, this work provides practical guidance for reliable and comparable evaluation in programming knowledge tracing.

\end{abstract}

\section{Introduction}\label{sec:intro}

Knowledge Tracing (KT) aims to model the temporal evolution of a learner's knowledge state by analyzing their historical interaction data. As programming education has become a cornerstone of the modern Computer Science curriculum, research into Programming Knowledge Tracing (PKT) has expanded rapidly. Unlike traditional KT domains such as mathematics, programming involves process-oriented problem-solving rather than single-correct-answer submissions. Learners engage in an iterative cycle of writing, executing, debugging, and refining code. Consequently, specialized PKT models have been developed to capture these sophisticated learning dynamics more precisely.

In response to these requirements, recent PKT approaches have adopted hybrid architectures that integrate feature modeling—reflecting the structural and semantic information of code—with sequential prediction models. For code representation, some PKT models employ Abstract Syntax Tree (AST)-based code extraction and attention-based encoding methods, such as code2vec~\cite{alon2019code2vec}, while others leverage Transformer-based pretrained models (e.g., CodeSage~\cite{zhang_codesage}). These code representations are then combined with Recurrent Neural Networks (RNNs) or Transformer architectures to track the learner’s evolving knowledge state.
These models have reported substantial performance gains over traditional DKT~\cite{piech2015_dkt} baselines and have become a prevalent modeling paradigm in recent PKT research. However, as model architectures grow increasingly complex, ensuring the reliability and reproducibility of reported performance has become a critical challenge.

In this study, we demonstrate that the empirical performance gains reported in representative PKT models~\cite{shi2022_codedkt,yu2024eckt} utilizing code2vec-based code representations can be highly sensitive to subtle implementation details and experimental design choices. Specifically, we identify implementation-level issues in attention-based code representation that directly affect performance estimates. Furthermore, we experimentally confirm that failing to align student attempt sequences by the time stamp results in temporal causality violations. This leads to significant data leakage, where future information inadvertently influences past predictions, fundamentally undermining the validity of sequence-based modeling and systematically overestimating model performance.

To systematically analyze these effects, we adopt a controlled evaluation protocol. We select optimal hyperparameters via grid search using a single designated fold ($fold_0$) and subsequently fix these parameters across all folds during cross-validation. This approach allows us to focus on how assignment-wise data characteristics influence model behavior. Additionally, we explore the impact of the maximum sequence length, a factor often arbitrarily fixed in prior studies despite significant variations in student interaction patterns.

By re-evaluating representative PKT models on the \textit{CodeWorkout} dataset using this protocol, we observe that the previously reported superiority of attention-augmented PK models is often diminished or even reversed under causally valid and consistent settings. Specifically, our results show that the performance advantage of the PK models is not consistent across tasks, and that traditional DKT can remain competitive—or even superior—under appropriate task-specific optimization. These findings suggest that many of the performance gains attributed to architectural innovations in prior work may have been overstated, arising from evaluation biases and inadequate baseline tuning rather than genuine improvements in predictive modeling.

The primary contributions of this work are as follows:
\begin{itemize}
    \item \textbf{Empirical Reliability Analysis}: We clarify how implementation details and sequence construction methods can systematically distort PKT performance metrics.
    \item \textbf{Controlled Evaluation Protocol}: We present a fair and reproducible benchmarking framework through a cross-validation process where data is partitioned \textbf{assignment-wise} into folds, and consistent hyperparameter settings are applied across all folds for each assignment.
    \item \textbf{Assignment and Sequence-level Insights}: We reveal critical but overlooked factors by analyzing the impact of assignment characteristics and maximum sequence length on PKT performance.
\end{itemize}

The source code for our experiments is available at \url{https://github.com/lesit/Reliable-PKT}.

\section{Related Work}

\subsection{Evolution of Knowledge Tracing}
Knowledge Tracing (KT) has evolved from Recurrent Neural Networks (RNNs) \cite{piech2015_dkt} to sophisticated attention-based architectures. For instance, SAINT~\cite{choi2020_saint} utilizes a Transformer encoder-decoder structure, while AKT~\cite{ghosh2020_akt} employs a monotonic attention mechanism with Rasch-based normalization. Recent models have introduced various mechanisms to enhance performance, such as removing noise through sparse attention (sparseKT)~\cite{huang2023sparseKT}, incorporating individual priors (AT-DKT)~\cite{liu2023_atdkt}, or modeling the forgetting process via linear biases (FoLiBiKT~\cite{im2023_folibikt} and extraKT~\cite{li2024_extrakt}). 

While these advancements have pushed the state-of-the-art in predictive AUC, recent audits suggest that architectural complexity can sometimes mask underlying experimental sensitivities~\cite{khajah2016deep}. Supporting this view, studies such as simpleKT~\cite{liu2023simplekt} and ReKT~\cite{shen2024_rekt} have demonstrated that streamlined architectures can achieve competitive or even superior results compared to more intricate models by focusing on core interaction patterns. Our work extends this critical perspective by investigating how these performance gains are influenced by fundamental experimental protocols in the programming domain.

\subsection{Programming Knowledge Tracing (PKT)}
PKT differs from traditional Knowledge Tracing (KT) in that it explicitly incorporates the iterative and process-oriented nature of coding activities. Early PKT models focused on incorporating program structure through Abstract Syntax Tree (AST)-based representations. 
Among them, Code-DKT~\cite{shi2022_codedkt} leverages code2vec~\cite{alon2019code2vec} to encode ASTs, combined with attention mechanisms and Long Short-Term Memory (LSTM) networks~\cite{hochreiter1997_lstm} to model students’ evolving knowledge states. 
Building on the Code-DKT architecture, more recent approaches such as ECKT~\cite{yu2024eckt} extend this line of work by integrating semantic-aware encoders and large language models to capture richer program semantics and problem-solving patterns.

\subsection{Benchmarking and Data Integrity}
The release of datasets like CodeWorkout~\cite{price_shi_codeworkout_2021} and ACcoding~\cite{dataset_accoding} has facilitated PKT research. However, evaluation protocols vary significantly across studies. Issues such as improper sequence alignment or arbitrary truncation of interaction histories can lead to temporal data leakage~\cite{hastie2009elements}, where future information inadvertently inflates performance estimates. This study identifies and rectifies these systemic biases to establish a more reliable benchmarking framework.

\section{Reliability Issues in Existing PKT Evaluations}
\label{sec:3_issues}

While recent PKT models have reported significant performance improvements, the reliability of these results depends heavily on the integrity of their underlying implementation and evaluation protocols. In this section, we conduct a systematic audit of the official open-source implementation~\cite{codedkt_github} of Code-DKT~\cite{shi2022_codedkt} to identify systemic evaluation biases. We find that this implementation contains critical structural flaws: (1) causal flaws in attention-based code representation ($CRect$, Section~\ref{subsec:3.1.temporal_leakage_by_att}), (2) temporal causality violations during sequence construction ($TAlign$, Section~\ref{subsec:3.2.temporal_leakage_by_timestamp}), and (3) performance distortions arising from hyperparameter sensitivity (Section~\ref{subsec:3.3.hparam_sensitivity}). Furthermore, we extend our comparative analysis to ECKT~\cite{yu2024eckt} specifically regarding causal rectification ($CRect$) to examine if these attention-based biases persist across related PKT architectures. By exposing these issues, we establish the necessity for the controlled evaluation protocol proposed in Section~\ref{sec:4.control_protocol}.

\subsection{Causal Invalidity in Attention-integrated Code Representation}
\label{subsec:3.1.temporal_leakage_by_att}

Recent PKT models have increasingly adopted hybrid architectures that integrate feature modeling to capture the semantic nuances of source code. A representative approach is Code-DKT~\cite{shi2022_codedkt}, which utilizes code2vec~\cite{alon2019code2vec} for code representation but augments its attention mechanism by incorporating student performance data ($x_t$) to assign weights to code paths. Specifically, the model represents student code $c_t$ as a set of $R$ leaf-to-leaf paths extracted from an Abstract Syntax Tree (AST). For each code path, the embeddings of the starting node, the path, and the ending node are concatenated with the student’s correctness vector $x_t$ to form a path representation $e_r$:
\begin{equation}
e_r = [e_{s,r}; e_{o,r}; e_{q,r}; x_t]
\end{equation}
The model employs a Score-Attended Path Selection mechanism where the attention weight $\alpha_r$ for each path is calculated using an attention matrix $W_a$:
\begin{equation}
\alpha_r = \frac{\exp(e_r W_a)}{\sum_{i=1}^{R} \exp(e_i W_a)}
\end{equation}

In our study, we executed the official source code~\cite{codedkt_github} provided by the authors and successfully replicated the empirical results reported in the original paper, including the performance gain of approximately 0.03--0.04 in AUC, a standard metric for binary classification performance, over the DKT~\cite{piech2015_dkt} baseline. During this replication process, we identified a critical structural flaw in the implementation: the Softmax operation in the attention layer is applied across the temporal dimension, specifically \texttt{dim=1}.

We argue that this configuration is causally invalid for the Knowledge Tracing task. Applying Softmax across the time-step dimension (\texttt{dim=1}) allows the model to normalize weights by considering the entire interaction sequence. This inherently introduces temporal data leakage, as the importance of code features at a specific time step $t$ is influenced by information from future states. Such a "look-ahead" mechanism violates the fundamental principle of Knowledge Tracing, which dictates that predictions must be conditioned strictly on current and historical data.

To maintain temporal causality and ensure the model correctly weights paths within a single code submission, the operation must be performed across the path dimension (\texttt{dim=2}). Our analysis demonstrates that the reported performance gains are not a result of architectural superiority but are consequences of this causal violation. By effectively bypassing temporal constraints, the model achieves an artificially inflated performance that does not reflect its true predictive capability.

\subsection{Temporal Causality Violation and Data Leakage}
\label{subsec:3.2.temporal_leakage_by_timestamp}

\begin{figure}[!t]
    \centering
    \includegraphics[width=1\linewidth]{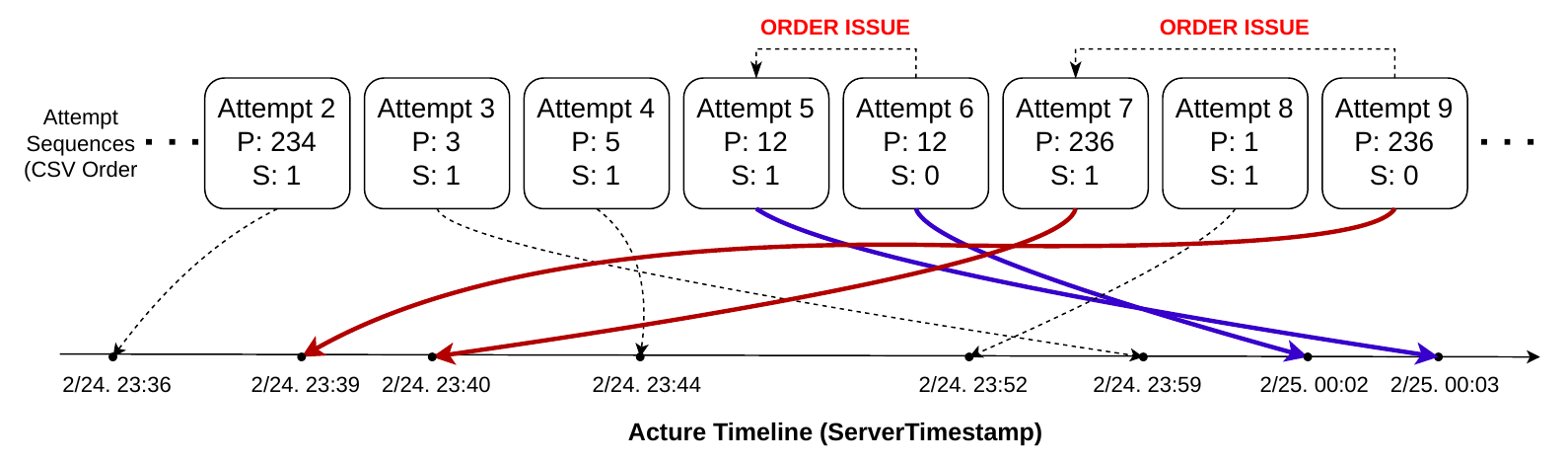}
    \caption{Chronological misalignment between dataset appearance order (CSV order) and actual server timestamps.}
    \label{fig_not_aligned}
\end{figure}

Knowledge Tracing (KT) models inherently assume that a learner’s future performance is conditioned only on past interactions. Therefore, preserving the strict chronological order of student interactions is a fundamental requirement for valid evaluation.

However, we identify a critical flaw in the official open-source implementation~\cite{codedkt_github} of Code-DKT~\cite{shi2022_codedkt}. Specifically, the codebase fails to perform chronological sorting by \texttt{ServerTimestamp} and instead aggregates interactions based on their raw appearance order. As illustrated in Figure~\ref{fig_not_aligned}, this leads to a direct violation of temporal causality. For instance, in the sequence for SubjectID 106, Attempt~6 (a failure) actually occurred before Attempt~5 (a success) in the actual timeline. When later attempts appear earlier in the input sequence, it introduces \emph{look-ahead bias} and explicit data leakage.

Formally, a causally valid model must operate on a sorted sequence $S'$ such that
\begin{equation}
P(a'_{t+1} \mid a'_1, \dots, a'_t)
\quad \text{where} \quad
\tau'_1 < \dots < \tau'_{t+1}.
\end{equation}
Our analysis reveals that this condition is frequently violated, particularly in programming contexts with rapid ``submit-and-fix'' cycles. Without strict sorting, the model can exploit future information unavailable at prediction time, leading to spurious learning signals and a systematic overestimation of performance that obscures the true latent learning process.

\subsection{Hyperparameter Sensitivity Across Models and Assignments}
\label{subsec:3.3.hparam_sensitivity}

Before introducing our assignment-wise cross-validation protocol, we examine the sensitivity of PKT models (Code-DKT~\cite{shi2022_codedkt}, ECKT~\cite{yu2024eckt}) performance to hyperparameter choices. Although hyperparameter tuning is often treated as a secondary detail, our analysis indicates that its impact is substantial and highly dependent on both the model architecture and the programming assignment.

We observe that optimal configurations vary not only across different PKT models, but also across assignments within the same model. Differences in assignment difficulty, interaction length, and response distributions induce distinct optimization landscapes, such that hyperparameters effective for one assignment may lead to degraded performance on another. Notably, this variability is observed consistently across both DKT and PKT models, suggesting that hyperparameter sensitivity is an inherent characteristic of programming interaction data rather than a byproduct of architectural complexity.

These findings imply that adopting a single global hyperparameter configuration can obscure true model behavior and bias comparative evaluations, motivating the need for a task-aware evaluation strategy developed in the following sections.

\section{Controlled Evaluation Protocol}
\label{sec:4.control_protocol}
To address the causal flaws and hyperparameter sensitivity issues identified in Section~\ref{sec:3_issues}, this study proposes three evaluation protocols that ensure causal integrity and experimental fairness.

\subsection{Implementation Rectification for Causal Integrity}
\label{subsec:rectification}


To correct the dimensionality error of the Softmax operation pointed out in Section~\ref{subsec:3.1.temporal_leakage_by_att}, this study changes the normalization axis of the attention mechanism from the time axis ($\text{dim}=1$) to the path axis ($\text{dim}=2$). This is to fundamentally prevent temporal data leakage, where predictions at a specific point in time refer to future sequence information. The weight $\alpha_r$ defined in Equation (2) is now normalized only across AST paths within a single code submission, thereby restoring the causal validity of the modeling.

\subsection{Chronological Sequence Alignment}
\label{subsec:alignment}

To address the sequence composition error discussed in Section~\ref{subsec:3.2.temporal_leakage_by_timestamp}, we enforce a strict ascending sort by the \texttt{ServerTimestamp} column as a preprocessing protocol before processing all student attempt records. This is to reconstruct the actual student learning trajectories without relying on the physical storage order of the dataset. This sorting process eliminates look-ahead bias, where future successful debugging attempts within a sequence influence past predictions.

\subsection{Assignment-wise Cross-Validation and Reproducibility}
\label{subsec:fixed_protocol}

To address the hyperparameter sensitivity identified in Section~\ref{subsec:3.3.hparam_sensitivity}, we adopt an assignment-wise tuning strategy where models are trained and evaluated independently for each task. 
For each assignment, we identify the optimal hyperparameter configuration $\theta^{*}$ through a grid search on a designated validation fold ($fold_0$). 
This configuration is then fixed and applied uniformly across all remaining folds to prevent overfitting and ensure that performance variations arise solely from data variability.

We enforce strict reproducibility by utilizing independent fixed random seeds for each experimental stage. 
A seed governs the initial training and test set separation, while a separate, distinct seed is employed to generate identical five-fold cross-validation splits. 
Furthermore, every model is initialized with a consistent seed to eliminate stochastic variance during training. 
This multi-layered protocol justifies task-specific tuning while preserving fairness and comparability across all evaluated assignments.

\subsection{Impact of Maximum Sequence Length}
\label{subsec:4.4.seq_length}

\vspace{-0.5em}
\begin{table}[hbt!]
    \centering
    \caption{95th percentile of attempt sequence lengths by assignment.}
    \label{tab:95th_percentiles}
    \renewcommand{\arraystretch}{1.1}
    \small
    \setlength{\tabcolsep}{12pt}
    \begin{tabular}{lccccc}
    \hline
    Assignment ID & 439 & 487 & 492 & 494 & 502 \\ 
    \hline 
    95th Percentile & 90 & 109 & 135 & 91 & 81 \\
    \hline 
    \end{tabular} 
\end{table}

In many PKT implementations, the maximum sequence length is treated primarily as a 
technical constraint for computational efficiency, as exemplified by Code-DKT, which 
adopts $L_{max}=50$~\cite{shi2022_codedkt}. However, programming education often involves 
extended sequences of iterative debugging, where aggressive truncation may remove 
important problem-solving context.

As shown in Table~\ref{tab:95th_percentiles}, the 95th percentile of student attempt 
sequence lengths consistently exceeds the commonly adopted limit of $L_{max}=50$, 
ranging from 81 to 135 across assignments. This observation indicates that a substantial portion of student interaction histories 
is not fully captured under the conventional setting. 
However, incorporating longer sequences does not necessarily guarantee improved modeling, 
as extended interaction histories may also introduce irrelevant or noisy information.

Motivated by this analysis, we explicitly compare two predefined settings, $L_{max}=50$ 
and $L_{max}=100$, to examine whether preserving longer interaction histories leads to 
measurable performance differences. The corresponding experimental results are reported 
in Section~\ref{subsec:6.4.seq_length}.

\section{Experimental Setup}

\subsection{Dataset and Sequence Analysis}
\label{subsec:dataset}
We utilize the \textit{CodeWorkout} dataset~\cite{price_shi_codeworkout_2021}, consisting of 69,627 interactions from 413 students across five assignments. To account for varying problem-solving durations, we analyzed the sequence length distribution for each assignment. 

\subsection{Evaluation Protocol and Hyperparameters}
\label{subsec:setup}
For each assignment, data is split into an 80\% training set and a 20\% test set, with the training set further partitioned into five folds. To ensure causal integrity and fair benchmarking among DKT~\cite{piech2015_dkt}, Code-DKT~\cite{shi2022_codedkt}, and ECKT~\cite{yu2024eckt}, we implement a controlled tuning procedure: 
\begin{enumerate}
    \item \textbf{Fold Configuration}: One fold is designated for validation while the other four are used for training. 
    \item  \textbf{Grid Search on $fold_0$}: We perform an exhaustive search exclusively on the first fold to identify the optimal configuration $\theta^*$. For DKT, the learning rate is tuned within $\{5e-5, 1e-4, 5e-4\}$. For Code-DKT and ECKT, we tune the embedding size $\in \{50, 100, 150, 300, 350\}$, $dropout \in \{0.1, 0.2, 0.3, 0.4, 0.5\}$, and learning rate $\in \{5e-5, 1e-4, 5e-4\}$. 
    \item  \textbf{Hyperparameter Fixation}: Once $\theta^*$ is determined, it is fixed across all subsequent folds to ensure that performance variations stem solely from data variability and to prevent hyperparameter-induced overfitting.
\end{enumerate}

\section{Results and Analysis}

This section presents a comprehensive re-evaluation of DKT~\cite{piech2015_dkt} and models (Code-DKT~\cite{shi2022_codedkt}, ECKT~\cite{yu2024eckt}), progressing from faithful reproduction to a more rigorous, task-aware evaluation protocol. 
In Section~\ref{subsec:6.1.rect}, we reproduce the original Code-DKT performance and compare it with results obtained after rectifying sequence alignment ($TAlign$) and the softmax dimension in the attention mechanism ($CRect$). 
Building on these observations, Sections \ref{subsec:6.2.opt} and \ref{subsec:6.3.model} adopt a more granular protocol based on assignment-wise cross-validation. For each model--task pair, hyperparameters are tuned exclusively on a designated validation fold ($fold_0$) and then fixed across all five folds to ensure fairness and reproducibility. All results are reported as the mean $\pm$ standard deviation across folds.

\subsection{Impact of Chronological Alignment and Causal Rectification}
\label{subsec:6.1.rect}

\vspace{-0.5em}
\begin{table}
    \caption{Impact of Chronological Alignment and Causal Rectification on Model Performance} 
    \label{tab:re-eval}
    \centering
    \renewcommand{\arraystretch}{1.1}
    \footnotesize
    \setlength{\tabcolsep}{2pt}
    \resizebox{\textwidth}{!}{%
	\begin{tabular}{lllllll}
	\hline
	Model & Setting & 439 & 487 & 492 & 494 & 502\\ 
	\hline 

	DKT & $-$ & $0.6768{\pm}0.02$ & $0.7359{\pm}0.01$ & $0.7650{\pm}0.02$ & $0.6838{\pm}0.02$ & $0.7469{\pm}0.01$ \\ 
	CodeDKT & $-$ & $0.7445{\pm}0.01$ & $0.7651{\pm}0.01$ & $0.8049{\pm}0.01$ & $0.7265{\pm}0.01$ & $0.7932{\pm}0.01$ \\ 
	\hline 
	*DKT & $TAlign$ & $0.6821{\pm}0.02$ & $0.7065{\pm}0.01$ & $0.7312{\pm}0.02$ & $0.7184{\pm}0.02$ & $0.7358{\pm}0.01$ \\ 
	CodeDKT & $TAlign$ & $0.7116{\pm}0.01$ & $0.7659{\pm}0.01$ & $0.7895{\pm}0.01$ & $0.7465{\pm}0.01$ & $0.7928{\pm}0.01$ \\ 
	*CodeDKT & $CRect$ & $0.6914{\pm}0.01$ & $0.7512{\pm}0.01$ & $0.7716{\pm}0.01$ & $0.7235{\pm}0.01$ & $0.7695{\pm}0.01$ \\ 

	\hline 
	\end{tabular} 
    }
\end{table} 

In this section, we examine the performance impact of a corrected evaluation pipeline. Specifically, we evaluate chronological alignment ($TAlign$) to DKT~\cite{piech2015_dkt}, while for PKT models (Code-DKT~\cite{shi2022_codedkt}, ECKT~\cite{yu2024eckt}), we evaluate the combined effect of $TAlign$ and causal rectification ($CRect$). Rather than seeking performance gains, this re-evaluation aims to establish a transparent baseline by eliminating potential data leakage and implementation oversights present in prior work.

The results in Table~\ref{tab:re-eval} show that applying $TAlign$ leads to inconsistent performance shifts. For instance, while DKT's AUC on dataset 439 slightly increases from $0.6768$ to $0.6821$, Code-DKT experiences a noticeable drop on several datasets (e.g., $0.7445 \to 0.7116$ on 439). This suggests that previous high scores might have been partially inflated by the inadvertent use of future information within non-chronological sequences.

Furthermore, the introduction of $CRect$ reveals that the original implementation of Code-DKT was sensitive to the dimensionality of the softmax operation. The adjustment of the attention mechanism ensures causal integrity, providing a more rigorous assessment of the model’s ability to predict future performance based solely on past interactions.

\vspace{-0.5em}
\begin{table}[hbt!]
    \centering
    \caption{Performance comparison (AUC) of DKT and PKT models (Code-DKT, ECKT) under task-specific hyperparameter optimization with chronologically aligned sequences. $TAlign$ denotes chronological sequence alignment, $CRect$ indicates causal rectification of the attention computation, and $CRect+$ further includes the $W_0$ component described in the original model specification. Standard deviations are omitted for brevity as they are consistently below 0.01, except for DKT on Assignment 502 ($\pm 0.04$).}
    \label{tab:tuning_rect}
    \renewcommand{\arraystretch}{1.1}
    \small
    \setlength{\tabcolsep}{8pt}
    \begin{tabular}{lllllll}
    \hline
	Model & Setting & 439 & 487 & 492 & 494 & 502\\ 
    \hline 

	*DKT & $TAlign$ & $\textbf{0.7480}$ & $0.7518$ & $0.7705$ & $0.7630$ & $0.7490$ \\ 
    \hline
	CodeDKT & $TAlign$ & $0.7254$ & $0.7672$ & $0.7905$ & $0.7507$ & $0.7967$ \\ 
	*CodeDKT & $CRect$ & $0.7120$ & $0.7604$ & $0.7864$ & $0.7082$ & $0.7768$ \\ 
	*CodeDKT & $CRect+$ & $0.7268$ & $0.7633$ & $0.7831$ & $0.7162$ & $0.7731$ \\ 
	*ECKT & $CRect$ & $0.7239$ & $\textbf{0.7762}$ & $\textbf{0.8050}$ & $0.7463$ & $0.7947$ \\ 
	*ECKT & $CRect+$ & $0.7364$ & $0.7588$ & $0.8017$ & $\textbf{0.7632}$ & $\textbf{0.7958}$ \\ 
    
    \hline 
    \end{tabular} 
\end{table} 

\subsection{Criticality of Task-specific Hyperparameter Optimization}
\label{subsec:6.2.opt}

Based on the corrected evaluation pipeline established in Section~\ref{subsec:6.1.rect}, Table~\ref{tab:tuning_rect} presents the performance of DKT~\cite{piech2015_dkt} and PKT models (Code-DKT~\cite{shi2022_codedkt}, \allowbreak ECKT~\cite{yu2024eckt}) when provided with task-specific hyperparameter optimization. While the experiments in Section~\ref{subsec:6.1.rect} adhered to the original studies' fixed settings (e.g., a 40-epoch limit) to isolate the impact of $TAlign$ and $CRect$, the results here reflect each model's true competitive capacity achieved through independent tuning and early stopping for each assignment. Our findings demonstrate that applying uniform training configurations across different models and tasks can lead to a systematic underestimation of performance, particularly for baseline architectures. Consequently, we establish that task-specific optimization is a prerequisite for any reliable comparative evaluation, ensuring that architectural advantages are assessed only after each model has reached its optimal state.

\begin{table}[hbt!]
    \centering
    \caption{Task-specific optimal learning rates and early-stopped epochs for DKT~\cite{piech2015_dkt}.}
    \label{tab:hp_dkt}
    \renewcommand{\arraystretch}{1.1}
    \small
    \setlength{\tabcolsep}{12pt}
    \begin{tabular}{ccc}
    \hline
	Assignment & Learning rate & Epoch \\ 
    \hline 
    439 & $0.0005$ & $42.4 \pm 21.7$ \\
    487 & $0.0005$ & $68.4 \pm 15.9$ \\
    492 & $0.0005$ & $46.2 \pm 08.4$ \\
    494 & $0.0005$ & $75.2 \pm 24.6$ \\
    502 & $0.0001$ & $46.0 \pm 30.2$ \\
    \hline 
    \end{tabular} 
\end{table} 

A particularly illustrative case is Assignment~439, where the AUC of DKT improves from $0.6821$ to $0.7480$ (approximately $+0.07$) solely through task-specific tuning. As summarized in Table~\ref{tab:hp_dkt}, the optimal training duration for DKT varies substantially across assignments, with mean early-stopped epochs ranging from $42.4$ to $75.2$. Notably, Assignment~494 requires nearly twice as many training epochs as the previously fixed setting, while Assignment~502 favors a smaller learning rate ($1e\!-\!4$) than other tasks.

These results indicate that previously reported performance gaps between DKT and more complex PKT models may stem, at least in part, from insufficient optimization of baseline models rather than intrinsic architectural advantages. Since hyperparameter sensitivity is jointly dependent on both the model architecture and the task context, task-specific optimization is not an optional refinement but a prerequisite for rigorous and fair comparative evaluation in programming knowledge tracing.

\subsection{Comparative Analysis of Rectified PKT Models}
\label{subsec:6.3.model}

We now examine the performance of DKT~\cite{piech2015_dkt} and PKT models (Code-DKT~\cite{shi2022_codedkt}, ECKT~\cite{yu2024eckt}) after correcting issues at the implementation level. Models marked with an asterisk ($*$) ensure causal validity, which is achieved through temporally aligned ($TAlign$) sequences for DKT, and by both $TAlign$ and a causally valid attention computation ($CRect$) for PKT models, with $CRect+$ further denoting the inclusion of the $W_0$ component.

As shown in Table~\ref{tab:tuning_rect}, rectifying the attention computation does not uniformly improve performance. In several assignments (e.g., 439 and 494), the rectified Code-DKT model underperforms the tuned DKT baseline despite utilizing code-level representations. This indicates that the inclusion of code embeddings, such as those from code2vec~\cite{alon2019code2vec}, does not consistently improve performance under causally valid and optimized settings.

Notably, our results for both Code-DKT and ECKT show that restoring the $W_0$ component ($CRect+$) yields only marginal changes in AUC across most assignments. This limited empirical impact suggests that the core predictive behavior is dominated by sequence modeling and task-specific dynamics rather than this specific projection weight.

In contrast, ECKT exhibits a more stable performance profile. Even without the $W_0$ component, the $CRect$ version of ECKT consistently matches or outperforms DKT, effectively narrowing the gap in assignments where Code-DKT struggles. These findings suggest that the structured integration of programming context in ECKT provides a robust advantage over simple feature concatenation. Overall, the effectiveness of PKT models depends on causally valid implementations and the alignment between model assumptions and task characteristics, rather than architectural complexity alone.
\subsection{Impact of Maximum Sequence Length}
\label{subsec:6.4.seq_length}

\vspace{-0.5em}
\begin{table}
	\caption{Performance comparison of DKT and PKT models (Code-DKT, ECKT) with maximum sequence length extended to 100. DKT results are reported under chronological alignment ($TAlign$), while PKT models are evaluated with causal rectification ($CRect$). $\Delta$ denotes the AUC variance relative to the baseline ($L=50$).}
    \label{tab:seq_comparison}
	\centering 
    \renewcommand{\arraystretch}{1.1}
    \setlength{\tabcolsep}{2pt}
    \resizebox{\textwidth}{!}{%
	\begin{tabular}{llccccc}
	\hline
	Model & Setting & 439 & 487 & 492 & 494 & 502\\ 

	\hline 
    \multirow{2}{*}{*DKT}
	 & $TAlign$ & $0.7147{\pm}0.04$ & $0.7640{\pm}0.00$ & $0.7677{\pm}0.02$ & $0.7193{\pm}0.02$ & $0.7608{\pm}0.01$ \\
	 & \small $\Delta$ & ${-0.0333}$ & $\textbf{+0.0122}$ & ${-0.0028}$ & ${-0.0437}$ & $\textbf{+0.0118}$ \\

	\hline 
    \multirow{2}{*}{*CodeDKT}
     & $CRect$ & $0.7308{\pm}0.00$ & $0.7434{\pm}0.01$ & $0.7754{\pm}0.01$ & $0.7088{\pm}0.01$ & $0.7604{\pm}0.01$ \\
	 & \small $\Delta$ & $\textbf{+0.0188}$ & ${-0.0170}$ & ${-0.0110}$ & $\textbf{+0.0006}$ & ${-0.0164}$ \\

	\hline 
    \multirow{2}{*}{*CodeDKT}
     & $CRect+$ & $0.7051{\pm}0.00$ & $0.7354{\pm}0.01$ & $0.7728{\pm}0.01$ & $0.7057{\pm}0.01$ & $0.7567{\pm}0.00$ \\
	 & \small $\Delta$ & ${-0.0217}$ & ${-0.0279}$ & ${-0.0103}$ & ${-0.0105}$ & ${-0.0164}$ \\

	\hline 
    \multirow{2}{*}{*ECKT}
     & $CRect$ & $0.7214{\pm}0.00$ & $0.7761{\pm}0.00$ & $0.7981{\pm}0.01$ & $0.7325{\pm}0.00$ & $0.7653{\pm}0.01$ \\
	 & \small $\Delta$ & ${-0.0025}$ & ${-0.0001}$ & ${-0.0069}$ & ${-0.0138}$ & ${-0.0294}$ \\

	\hline 
    \multirow{2}{*}{*ECKT}
     & $CRect+$ & $0.7316{\pm}0.00$ & $0.7709{\pm}0.01$ & $0.7908{\pm}0.01$ & $0.7205{\pm}0.01$ & $0.7463{\pm}0.01$ \\
	 & \small $\Delta$ & ${-0.0048}$ & $\textbf{+0.0121}$ & ${-0.0109}$ & ${-0.0427}$ & ${-0.0495}$ \\ 
    
	\hline 
	\end{tabular} 
    }
\end{table} 

Table~\ref{tab:seq_comparison} reports the performance differences between the conventional 
setting ($L_{max}=50$) and the extended context setting ($L_{max}=100$). Despite the strong 
motivation for using longer sequences discussed in Section~\ref{subsec:4.4.seq_length}, extending 
the maximum sequence length does not lead to consistent performance improvements.

Across both DKT~\cite{piech2015_dkt} and PKT models (Code-DKT~\cite{shi2022_codedkt}, ECKT~\cite{yu2024eckt}), most assignments exhibit either marginal changes or clear 
performance degradation when $L_{max}$ is increased to 100. Performance gains are sporadic 
and limited in magnitude, whereas negative $\Delta$ values appear more frequently and 
consistently across models and datasets.

These results suggest that, although longer interaction histories are structurally 
well-motivated in programming education, simply increasing the context window is insufficient 
to improve knowledge tracing performance. Instead, extended sequences may introduce additional 
irrelevant or weakly informative interactions, offsetting the potential benefits of longer 
temporal context.

\section{Discussion and Limitations}
\label{sec:discussion}
Our results indicate that the reported performance gains of PKT models (Code-DKT~\cite{shi2022_codedkt}, ECKT~\cite{yu2024eckt}) are highly sensitive to evaluation and optimization choices. In particular, task-specific hyperparameter tuning reveals that baseline models such as DKT~\cite{piech2015_dkt} can be substantially underestimated under standardized settings, leading to potentially biased architectural comparisons. Moreover, the inconsistent gains of Code-DKT across assignments suggest that the effectiveness of code-level representations is task-dependent rather than universal. This study is limited to a specific dataset (CodeWorkout~\cite{price_shi_codeworkout_2021}) and a subset of PKT models; extending the proposed rectified evaluation protocol to broader domains and architectures remains an important direction for future work.

\section{Conclusion}
\label{sec:conclusion}
In this paper, we have re-examined the performance of PKT models (Code-DKT~\cite{shi2022_codedkt}, ECKT~\cite{yu2024eckt}) by identifying and rectifying critical biases in the evaluation pipeline. Through extensive experiments, we demonstrated that chronological alignment and causal rectification are essential for valid model assessment. Furthermore, we showed that per-assignment hyperparameter optimization is a prerequisite for fair comparison, as it reveals the true competitive capacity of baseline models like DKT~\cite{piech2015_dkt}. By providing a rectified evaluation protocol, our work offers a more rigorous foundation for measuring progress in programming knowledge tracing. We believe that the insights and standardized procedures established in this study will guide future research toward more reliable and transparent evaluations in the field of intelligent tutoring systems. Collectively, this work contributes (i) an empirical analysis of evaluation biases in programming knowledge tracing, (ii) a controlled and reproducible evaluation protocol, and (iii) task- and sequence-level insights that clarify when architectural complexity translates into genuine performance gains.

\section*{Acknowledgments}
This work was supported by the National Research Foundation of Korea(NRF) grant funded by the Korea government(Ministry of Science and ICT)(No.RS2025-16064585).

%
%
%
\bibliographystyle{plain}
\bibliography{reference}
%


\end{document}